\documentclass[10pt, conference]{IEEEtran}
\usepackage{config/ieee-paper}
\usepackage{tabularx}
\usepackage{booktabs}
\usepackage{tikz}
\usepackage{multirow}
\usepackage{placeins} 
\usetikzlibrary{positioning,shapes,arrows}
\def\BibTeX{{\rm B\kern-.05em{\sc i\kern-.025em b}\kern-.08em
    T\kern-.1667em\lower.7ex\hbox{E}\kern-.125emX}}

\makeatletter
\newcommand{\linebreakand}{%
  \end{@IEEEauthorhalign}
  \hfill\mbox{}\par
  \mbox{}\hfill\begin{@IEEEauthorhalign}
}
\makeatother

\begin{document}

\title{CBF-AFA:  Chunk-Based Multi-SSL Fusion for Automatic Fluency Assessment
}

\author{
    \IEEEauthorblockN{Papa Séga WADE}
    \IEEEauthorblockA{
        \textit{Orange Innovation, Châtillon, France} \\
        \textit{IMT-Atlantique, Lab-STICC }\\
        UMR CNRS 6285 Brest, France \\
        papasega.wade@orange.com
    }
    \and
    \IEEEauthorblockN{Mihai ANDRIES}
    \IEEEauthorblockA{
        \textit{IMT-Atlantique, Lab-STICC} \\
        \textit{UMR CNRS 6285, F-29238}\\
        Brest, France \\
        mihai.andries@imt-atlantique.fr
    }
    \and
    \IEEEauthorblockN{Ioannis KANELLOS}
    \IEEEauthorblockA{
        \textit{IMT-Atlantique, Lab-STICC} \\
        \textit{UMR CNRS 6285, F-29238}\\
        Brest, France \\
        ioannis.kanellos@imt-atlantique.fr
    }
    \linebreakand 
    \IEEEauthorblockN{Thierry MOUDENC}
    \IEEEauthorblockA{
        \textit{Orange Innovation} \\
        Lannion, France \\
        thierry.moudenc@orange.com
    } 
}

\maketitle

\begin{abstract}
Automatic fluency assessment (AFA) remains challenging, particularly in capturing speech rhythm, pauses, and disfluencies in non-native speakers. We introduce a chunk-based approach integrating Self-supervised learning (SSL) models—Wav2Vec2, HuBERT, and WavLM—selected for their complementary strengths in phonetic, prosodic, and noisy speech modeling, with a hierarchical CNN-BiLSTM framework. Speech is segmented into breath-group chunks using Silero voice activity detection (Silero-VAD), enabling fine-grained temporal analysis while mitigating over-segmentation artifacts. SSL embeddings are fused via a learnable weighted mechanism, balancing acoustic and linguistic features, and enriched with chunk-level fluency markers (e.g., speech rate, pause durations, n-gram repetitions). The CNN-BiLSTM captures local and long-term dependencies across chunks. Evaluated on Avalinguo and Speechocean762, our approach improves F1-score by 2.8 and Pearson correlation by 6.2 points over single SSL baselines on Speechocean762, with gains of 4.2 F1-score and 4.0 Pearson points on Avalinguo, surpassing Pyannote.audio-based segmentation baselines. These findings highlight chunk-based multi-SSL fusion for robust fluency evaluation, though future work should explore generalization to dialects with irregular prosody.

\end{abstract}


\section{Introduction}

Linguistic fluency is a multidimensional construct encompassing speech rate, pauses, syntactic complexity, and coherence, influenced by both linguistic (grammar, vocabulary) and non-linguistic (hesitations, pauses) factors. Assessments typically rely on subjective ratings or objective acoustic metrics, each with inherent limitations.

Early AFA methods used handcrafted acoustic features (e.g., speech rate, syllable counts, pauses) combined with statistical and deep learning (DL) models, such as (SVMs, LSTMs)~\cite{10289791},\cite{Deng2020},\cite{Deng2021},\cite{Zhang2021}. However these approaches required extensive feature engineering and large annotated datasets. While DL reduced manual feature extraction, data scarcity remains a key challenge.

SSL models Wav2Vec2~\cite{baevski2020wav2vec}, HuBERT~\cite{hsu2021hubert}, and WavLM~\cite{chen2022wavlm} learn speech representations directly from raw audio, each excelling in distinct aspects: Wav2Vec2 captures phonetic content, HuBERT models prosodic structure, and WavLM enhances robustness in noisy speech~\cite{chen2022wavlm}. These models complement each other by providing distinct fluency cues~\cite{chen2022wavlm, shekar23b_interspeech, Disentanglingarticle}. However, most AFA studies rely on full-utterance or frame-level analysis, which may overlook fine-grained disfluencies~\cite{defino22_interspeech}. 

We introduce a chunk-based AFA framework that integrates Silero Voice Activity Detection (Silero-VAD)~\cite{silero2021} for breath-group segmentation, multi-SSL fusion (Wav2Vec2, HuBERT, WavLM), and explicit fluency markers (speech rate, pause frequency). A CNN-BiLSTM architecture captures both local and global dependencies across speech chunks. Our key hypotheses are:
(1) Chunking enables more fine-grained disfluency analysis than full-utterance processing,
(2) Multi-SSL fusion enhances representational diversity, 
(3) Explicit fluency features complement SSL embeddings.  

Our main contributions are: 
\begin{itemize} 
\item A novel breath-group chunking approach for AFA,  
\item A learnable SSL fusion mechanism,  
\item An end-to-end CNN-BiLSTM framework integrating SSL embeddings and fluency markers.  
\end{itemize}
We validate our approach on Avalinguo~\cite{b8avalinguo} (1,388 conversational speech samples) and Speechocean762~\cite{zhang2021speechocean762} (5,000 scripted utterances), covering a range of fluency levels. The remainder of this paper is organized as follows: related works are discussed in Section~\ref{sec:relatedwork}, methodology in Section~\ref{sec:methodology}, experiments and results in Section~\ref{sec:experiment}, and conclusion in Section~\ref{sec:conclusion}.

\section{Related Work}
\label{sec:relatedwork}
AFA has evolved from relying on handcrafted acoustic features---such as speech rate and pause duration---to modern DL and SSL approaches \cite{Deng2020, Zhang2021}. Traditional systems often used Random Forests or Support Vector Machines trained on manually engineered features. However, they remained constrained by data scarcity and feature-selection bias. DL architectures, such as Bidirectional LSTMs, later improved temporal modeling \cite{10289791}, but their effectiveness remained contingent on large annotated datasets.

Recent SSL methods, such as Wav2Vec2 \cite{baevski2020wav2vec}, HuBERT \cite{hsu2021hubert}, and WavLM \cite{chen2022wavlm}, learn representations directly from raw audio, reducing reliance on transcripts. These have enabled ASR-free (Automatic Speech Recognition) fluency scoring \cite{liu2023asr} and pronunciation evaluation \cite{kim22k_interspeech}. However, fluency encompasses more than segmental aspects; factors such as speech flow and pause patterns play a crucial role in perceived fluency \cite{defino22_interspeech}. Studies leveraging SSL embeddings have shown promising results on non-native speech corpora such as Avalinguo and Speechocean762 \cite{10289791,b8avalinguo,chen2022wavlm}. However, many still process entire audio or simple frame-level splits, limiting fine-grained disfluency analysis.

Parallel research on speaker diarization and related tasks has highlighted the benefits of segmenting long audio into smaller chunks using frameworks like \texttt{pyannote.audio} \cite{bredin2020pyannote} or Silero-VAD modules such as Silero. However, the full potential of chunk-based segmentation for AFA remains underexplored. Our work addresses this gap by segmenting audio into breath-group chunks, integrating multiple SSL embeddings (Wav2Vec2, HuBERT, WavLM) via a learnable weighted fusion, and incorporating fluency markers (e.g., speech rate, pause frequency) within a hierarchical CNN-BiLSTM model. This approach combines fine-grained analysis of local disfluencies with robust global modeling, introducing a novel framework for AFA.

\section{Automated Fluency Assessment methodology}
\label{sec:methodology}

Our proposed end-to-end methodology for AFA combines chunk-based segmentation of audio, self-supervised learning (SSL) embeddings fused via a learnable weighted-sum, and additional fluency-related features (e.g., speech rate, pauses) within a CNN-BiLSTM classification architecture. The approach comprises the following steps:
\begin{enumerate}
    \item Silero-VAD and Breath-Group Chunking
    \item Fine-Tuning SSL \& Feature Extraction
    \item Weighted Fusion of SSL Embeddings
    \item Fluency Feature Computation
    \item CNN-BiLSTM Classification
\end{enumerate}
We detail each step below.

\subsection{Step 1: Silero-VAD and Breath-Group Chunking}
\label{sec:vad_chunking}

We first apply a Silero-VAD~\cite{silero2021} module to identify segments of active speech and discard irrelevant portions (e.g., long silences, background noise). Although utterances are pre-segmented, this additional VAD step enables finer prosodic chunking by identifying intra-utterance breath pauses. Specifically, we employ Silero-VAD, which generates start and end timestamps 
\begin{equation}
\mathcal{T} = \{(t_1^\text{start}, t_1^\text{end}), \dots, (t_N^\text{start}, t_N^\text{end})\}
\end{equation}
for detected speech regions in the input signal $\mathbf{x}$ as \newline $\mathcal{T} = \text{Silero-VAD}(\mathbf{x})$. Each VAD-segmented region is then further subdivided into \emph{breath-group chunks}. A breath group typically ends where the speaker takes a noticeable pause or breath, helping us capture natural prosodic boundaries. Practically, this can be done by detecting within-segment silences of duration above a threshold $\delta$. Let 
$\mathbf{x}_{\ell}$ denote the $\ell$-th speech region from Silero-VAD; 
breath-group chunking yields as $\mathbf{c}_i = \mathbf{x}_{\ell}[\text{start}_i : \text{end}_i] \, \ \text{with} \quad i=1, \dots, M$, where $\text{start}_i$ and $\text{end}_i$ mark the breath-group boundaries within that Silero-VAD segment. This multi-stage segmentation ensures that each chunk is both speech-only (thanks to Silero-VAD) and aligned with a natural prosodic boundary (breath group).

\subsection{Step2: Fine-Tuning SSL \& Feature Extraction}
\subsubsection{Fine-Tuned SSL Models for Fluency Embedding}
\label{sec:ssl_models }
To capture robust and nuanced acoustic features, we leverage three pre-trained SSL models — \emph{wav2vec2-large-960h-lv60-self}, \emph{hubert-large-ls960-ft }, and \emph{wavlm-large} — each fine-tuned on our fluency dataset using a Connectionist Temporal Classification (CTC) loss. This fine-tuning, on the training data, aligns audio sequences with fluency labels without requiring manual alignment, enabling the models to learn task-specific features while preserving their general speech representation capabilities.

\begin{itemize}
    \item \textbf{Wav2Vec2}: Learns contextualized representations directly from raw audio through a contrastive learning objective, excelling in capturing fine-grained phonetic details and temporal dynamics.
    \item \textbf{HuBERT}: Utilizes masked prediction of clustered acoustic units, emphasizing prosodic and structural patterns in speech, which are critical for modeling fluency.
    \item \textbf{WavLM}: Enhances robustness by jointly modeling speech content and speaker variations, making it particularly effective in diverse or noisy speaking conditions.
\end{itemize}
The embeddings from these models are fused to create a comprehensive representation that encapsulates phonetic, prosodic, and rhythmic aspects of non-native speech.

\subsubsection{SSL Feature Extraction}
\label{sec:ssl_extraction}

For each chunk $\mathbf{c}_i \in \mathbb{R}^{L}$, we extract high-level acoustic embeddings from three SSL models.
Formally, each model $\Phi_m$ (where $m \in \{\text{wav2vec}, \text{hubert}, \text{wavlm}\}$) outputs a frame-level representation:
\begin{equation}
\mathbf{F}_{m}^{(i)} = \Phi_m(\mathbf{c}_i) \in \mathbb{R}^{T_i \times d_m}
\end{equation}
where $T_i$ is the number of frames for chunk $i$ and $d_m$ is the model’s embedding dimension. We then apply a mean-pooling operation over time to obtain a fixed-dimensional vector.
\subsection{Step 3: Weighted Fusion of SSL Embeddings}
\label{sec:weighted_fusion}

To optimally exploit the complementary strengths of Wav2Vec2, HuBERT, and WavLM, we combine their chunk-level embeddings via a learnable weighted-sum mechanism (referred to as $\text{SSLFusion}$):
\begin{equation}
\mathbf{f}_{\text{fused}}^{(i)} 
= \alpha_1 \,\mathbf{f}_{\text{wav2vec}}^{(i)} 
+ \alpha_2 \,\mathbf{f}_{\text{hubert}}^{(i)} 
+ \alpha_3 \,\mathbf{f}_{\text{wavlm}}^{(i)}
\end{equation}
where $\alpha_1, \alpha_2, \alpha_3 \ge 0$ and $\sum_{j=1}^{3} \alpha_j = 1$. These fusion weights $\{\alpha_j\}$ are learned jointly with the downstream classification objective, allowing the model to automatically determine the relative importance of each SSL source. The fused embedding has dimension $ d=\max(d_{\text{wav2vec}}, d_{\text{hubert}}, d_{\text{wavlm}})$; $d=1024$.

\subsection{Step 4: Fluency Feature Computation}
\label{sec:fluency_feats}

In addition to the fused SSL embedding, we compute chunk-level fluency markers that capture disfluencies and temporal aspects of speech. Common metrics include:
\begin{itemize}
    \item $\text{SpeechRate}_i$: number of words per second in chunk $i$.
    \item $\text{PauseDuration}_i$: average silence length surrounding chunk $i$.
    \item $\text{ArticulationRate}_i$: syllables per second of active speech $i$.
    \item $\text{NgramRepetition}_i$: measure of repeated n-grams from transcriptions.
\end{itemize}
Let $\mathbf{m}_i \in \mathbb{R}^{k}$ gather these fluency markers for chunk $i$. We then concatenate $\mathbf{m}_i$ with the fused SSL embedding $\mathbf{f}_{\text{fused}}^{(i)}$:
\begin{equation}
\mathbf{h}_i = \text{Concat}\bigl(\mathbf{f}_{\text{fused}}^{(i)}, \mathbf{m}_i\bigr) \in \mathbb{R}^{d_{\text{fused}}+k}
\end{equation}
This enriched chunk representation $\mathbf{h}_i$ encodes both acoustic/linguistic cues from SSL and explicit fluency metrics.
\subsection{Step 5: CNN-BiLSTM Classification}
\label{sec:cnn_bilstm}
Each utterance, segmented into $M$ breath-group chunks, is processed by a CNN-BiLSTM model to classify fluency levels (\texttt{Low}, \texttt{Medium} or \texttt{Intermediate}, \texttt{High}). This architecture conceptually resembles dual-path modeling in DPRNN~\cite{dual_path}.

A 1D convolutional layer with $128$ filters ($kernel=3, stride=1$) extracts local acoustic patterns from each chunk, capturing diverse temporal features. Its outputs are passed to a 2-layer Bidirectional LSTM (BiLSTM) ($hidden\_size=256$), which models dependencies across chunks by processing the sequence in both forward and backward directions. The final BiLSTM outputs are aggregated via mean-pooling, passed through a fully connected layer, and classified using a softmax function. Dropout ($p=0.3$) is applied to prevent overfitting with a learning rate of 1e-4.

\section{Experiment}
\label{sec:experiment}
Our experimental setup aims to validate the effectiveness of chunk-based segmentation and multi-SSL fusion for AFA. We first describe the datasets and preprocessing steps, followed by a study on the optimal segmentation threshold. Finally, we present the results of our comparative experiments. 
\subsection{Datasets}
We evaluate our approach on two publicly available non-native English corpora, summarized in Table~\ref{tab:datasets_overview}. Each utterance is labeled for fluency based on human annotations.
\subsubsection{Avalinguo Audio Dataset \cite{b8avalinguo}}
Avalinguo consists of 1424 audio recordings from non-native English speakers, labeled with \emph{Low}, \emph{Intermediate}, or \emph{High} fluency. After excluding 36 files with minimal speech, 1388 recordings remain, sampled at 16\,kHz and spanning approximately 2~hours of conversational speech. The speech regions range from 0.5\,s to 4.9\,s depending on the number of words spoken. We apply 5-fold cross-validation to ensure robust evaluation. A normalized version, including transcriptions and statistical analyses, is available online.\footnote{\url{https://huggingface.co/datasets/papasega/Avalinguo-Audio-Dataset-splitted}}


\subsubsection{Speechocean762 \cite{zhang2021speechocean762}}
Speechocean762 contains 5000 English utterances from 250 non-native speakers. Each speaker reads 20 sentences, and each utterance is rated by five human experts on a 0--10 fluency scale. Following a total duration of about 6\,hours (2.88\,h for training and 2.69\,h for testing). For consistency, we consolidate the original 0--10 ratings into three categories: \emph{Low\_fluency} (0--5), \emph{Medium\_fluency} (6--7), and \emph{High\_fluency} (8--10). The processed dataset, along with our fluency annotations, is publicly accessible.\footnote{\url{https://huggingface.co/datasets/papasega/speechocean762_fluency}}


\begin{table}[htbp]
\caption{Overview of the Avalinguo and Speechocean762 datasets.}
\label{tab:datasets_overview}
\centering
\begin{tabular}{|l|c|c|c|}
\hline
\textbf{Dataset} & \textbf{Subset} & \textbf{\#Segments} & \textbf{Duration} \\ 
\hline
\multirow{2}{*}{Speechocean762} & Train & 2500 & 2.88\,h \\
                                & Test  & 2500 & 2.69\,h \\
\hline
Avalinguo & Train/Test & 1388 & $\approx 2$\,h \\
\hline
\end{tabular}
\end{table}

\vspace{1mm}
\noindent

To ensure consistent input across models, all audio files are down sampled to $16 \ \text{kHz}$ and normalized for amplitude variations.
\subsection{Optimal chunking threshold analysis}
The segmentation threshold $\delta = $ 300 \text{ms} was empirically determined as optimal. F1-score and Pearson Correlation Coefficient (PCC) comparisons across 200, 250, 300, and 350 ms show that 300 ms yields the highest performance on both datasets. Shorter thresholds ($\le250 ms$) over-fragment speech, while longer ones ($\ge350 ms$) introduce noise.

Results confirm that 300 ms best preserves fluency cues, aligning with natural prosodic transitions while optimizing classification accuracy and correlation with human ratings.

\subsection{Baseline Comparisons and Ablation Study}
To assess the effectiveness of our chunk-based multi-SSL framework, we compare it against two baselines: (i) a frame-level segmentation method based on Pyannote.audio, and (ii) individual SSL models without chunking or fusion. An ablation study is also conducted to quantify the impact of chunking, fusion, and fluency-specific features.
\subsubsection{Baseline Comparisons}
Table~\ref{tab:baseline_results} summarizes the results obtained for both datasets.
\begin{table*}[h]
\caption{Comparison of baseline models and our proposed approach across Speechocean762 and Avalinguo datasets.}
\centering
\begin{tabular}{|l|c|c|c|c|}
\hline
\textbf{Model} & \multicolumn{2}{c|}{\textbf{Speechocean762}} & \multicolumn{2}{c|}{\textbf{Avalinguo}} \\
\cline{2-5}
 & \textbf{F1-score} & \textbf{PCC} & \textbf{F1-score} & \textbf{PCC} \\
\hline
Pyannote.audio Baseline & 0.759 & 0.676 & 0.857 & 0.825 \\
\hline
Best Single SSL Model (wavLM) & 0.802 & 0.749 & 0.931 & 0.926 \\
\hline
\citeauthor{kim22k_interspeech}
(Interspeech 2022) 
& \textbf{--} & {0.78} & \textbf{--}& \textbf{--} 
\\
\hline
\citeauthor{10289791} 
(EUSIPCO 2023) & {0.740} & \textbf{--} & {0.950} & \textbf{--} 
\\
\hline
\hline
\citeauthor{liu2023asr}
(ICASSP 2023) 
& \textbf{--} & {0.795} & \textbf{--}& \textbf{--} 
\\
\hline
\hline
\textbf{Fusion + Silero-VAD Chunking (Ours)} & \textbf{0.825} & \textbf{0.796} & \textbf{0.969} & \textbf{0.963} \\
\hline
\end{tabular}
\label{tab:baseline_results}
\end{table*}

\paragraph*{Observations:}
\begin{enumerate}
    \item Pyannote.audio baseline provides reasonable performance, particularly on Avalinguo, but is outperformed by all SSL-based models.
    \item Among the individual SSL models, WavLM achieves the best results, suggesting it captures fluency-relevant acoustic features more effectively.
    \item Our Fusion + Silero-VAD Chunking approach achieves the highest performance across both datasets, demonstrating the benefit of:
    \begin{itemize}
        \item Breath-group chunking, which enables better segmentation of fluency patterns.
        \item Multi-SSL fusion, which improves the model’s representational power.
    \end{itemize}
\end{enumerate}

\subsubsection{Ablation Study}
To evaluate the contribution of each component, we conduct an ablation study by removing key elements from the full model. WavLM outperforms other SSL models in both datasets. Fluency speech rate correlates best with WavLM, confirming its effectiveness in temporal modeling. Fusion of all three SSL models improves performance, leveraging complementary fluency cues. Silero-VAD-based chunking further boosts results, capturing speech rhythm and pauses. Multi-SSL fusion and chunking are crucial for accurate AFA.

\subsubsection{Voice Quality Analysis and Fluency Patterns}
\begin{figure}[h]
    \centering
    \includegraphics[width=0.40\textwidth]{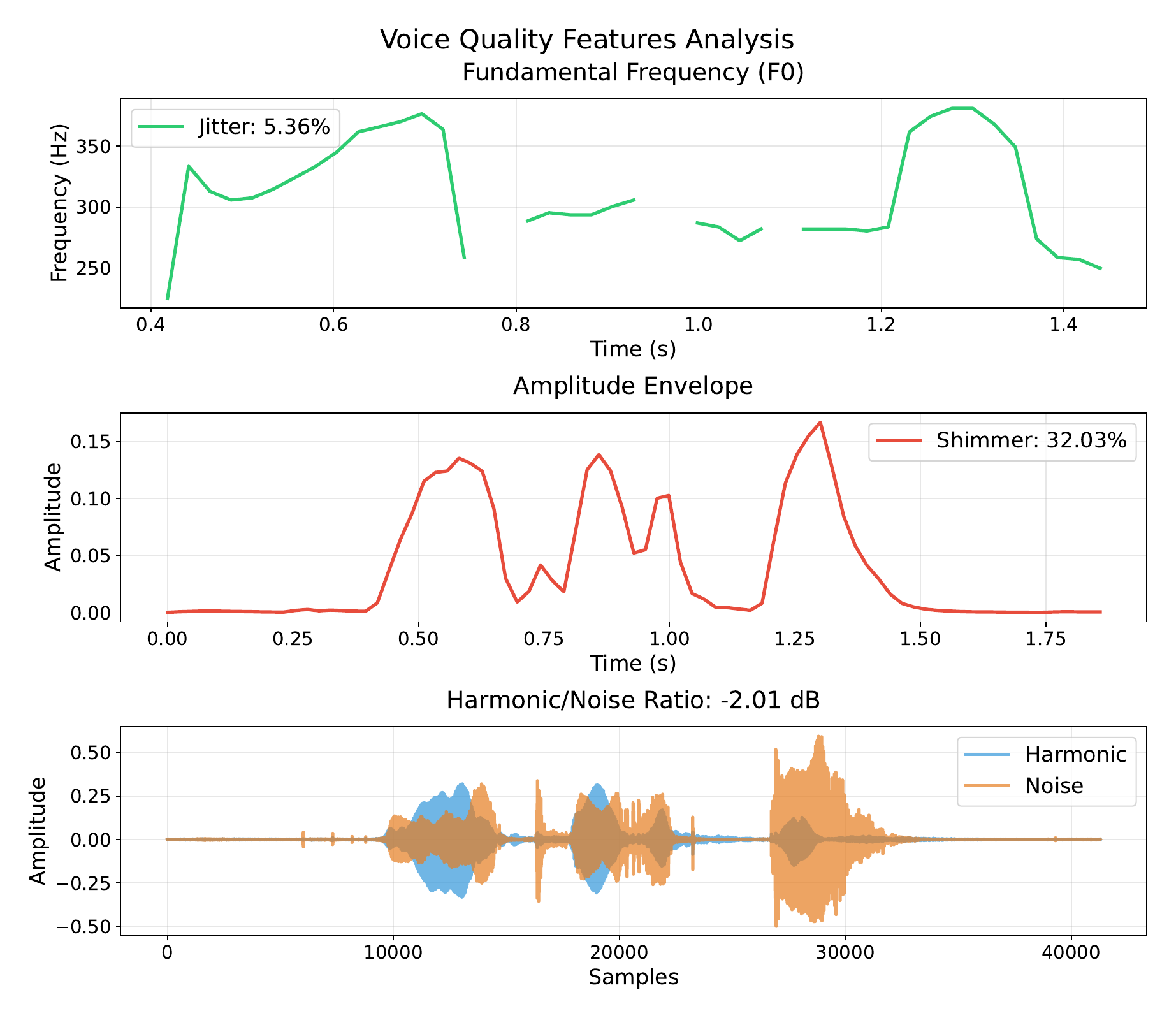}
    \caption{Voice Quality Analysis (F0, shimmer, HNR).}
    \label{fig:voice_quality}
\end{figure}

Beyond SSL embeddings, we analyze voice quality features, fundamental frequency (F0), shimmer, and harmonic-to-noise ratio (HNR), to examine their correlation with fluency. Figure~\ref{fig:voice_quality} shows that, in an audio sample, high shimmer (32.03\%) and negative HNR (-2.01 dB) suggest vocal instability and spectral noise, common in non-native or hesitant speech. F0 variations further highlight prosodic irregularities.

These findings support the idea that voice quality metrics complement SSL-based fluency assessment, providing additional insights into speech rhythm and stability.

\subsubsection{t-SNE Visualization of SSL Embeddings}
\begin{figure}[h]
    \centering
    \includegraphics[width=0.46\textwidth]{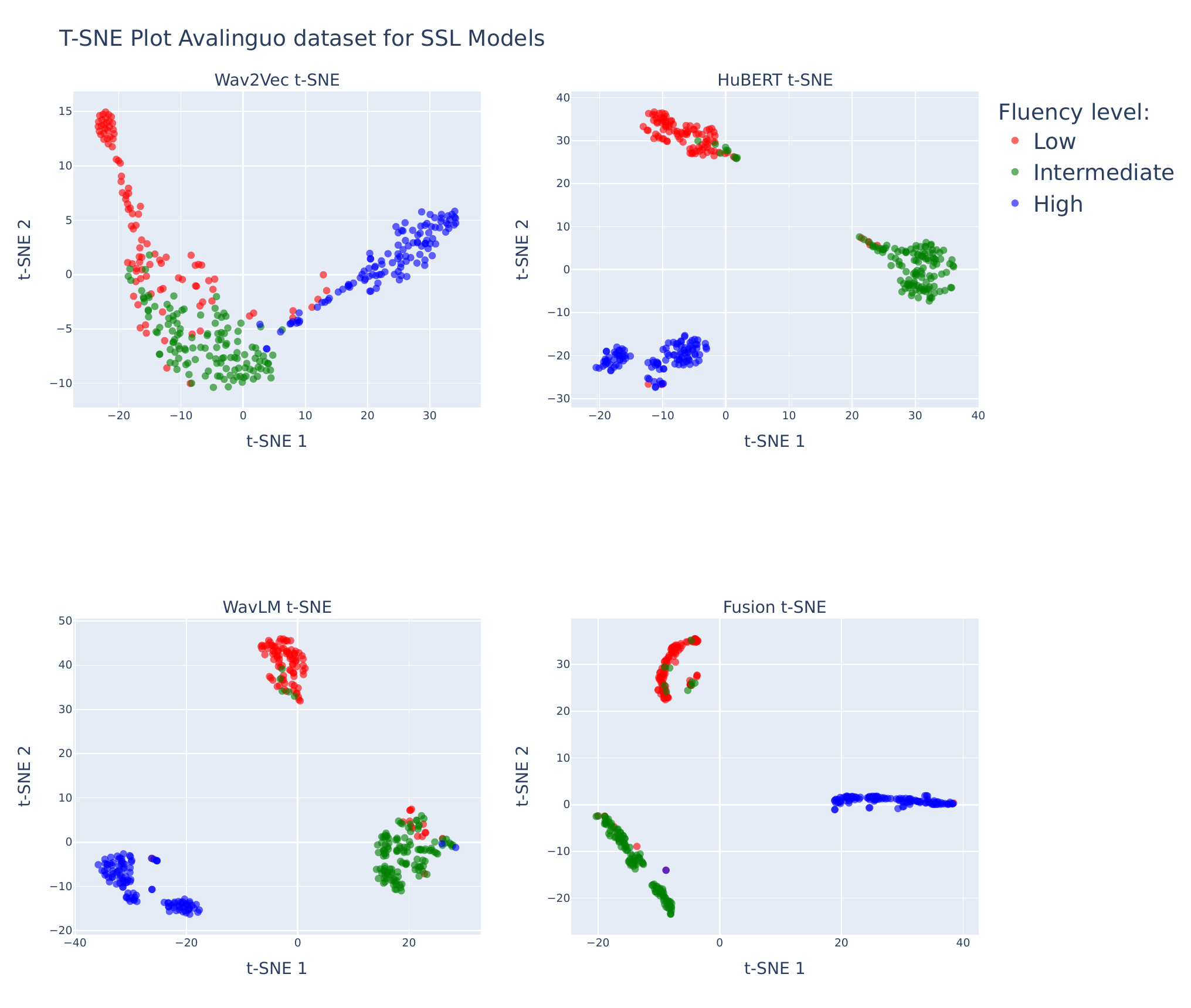}
   \caption{t-SNE Visualization of SSL embeddings (Avalinguo)}
    \label{fig:tsnet_avalinguo}
\end{figure}

\begin{figure}[h]
    \centering
    \includegraphics[width=0.46\textwidth]{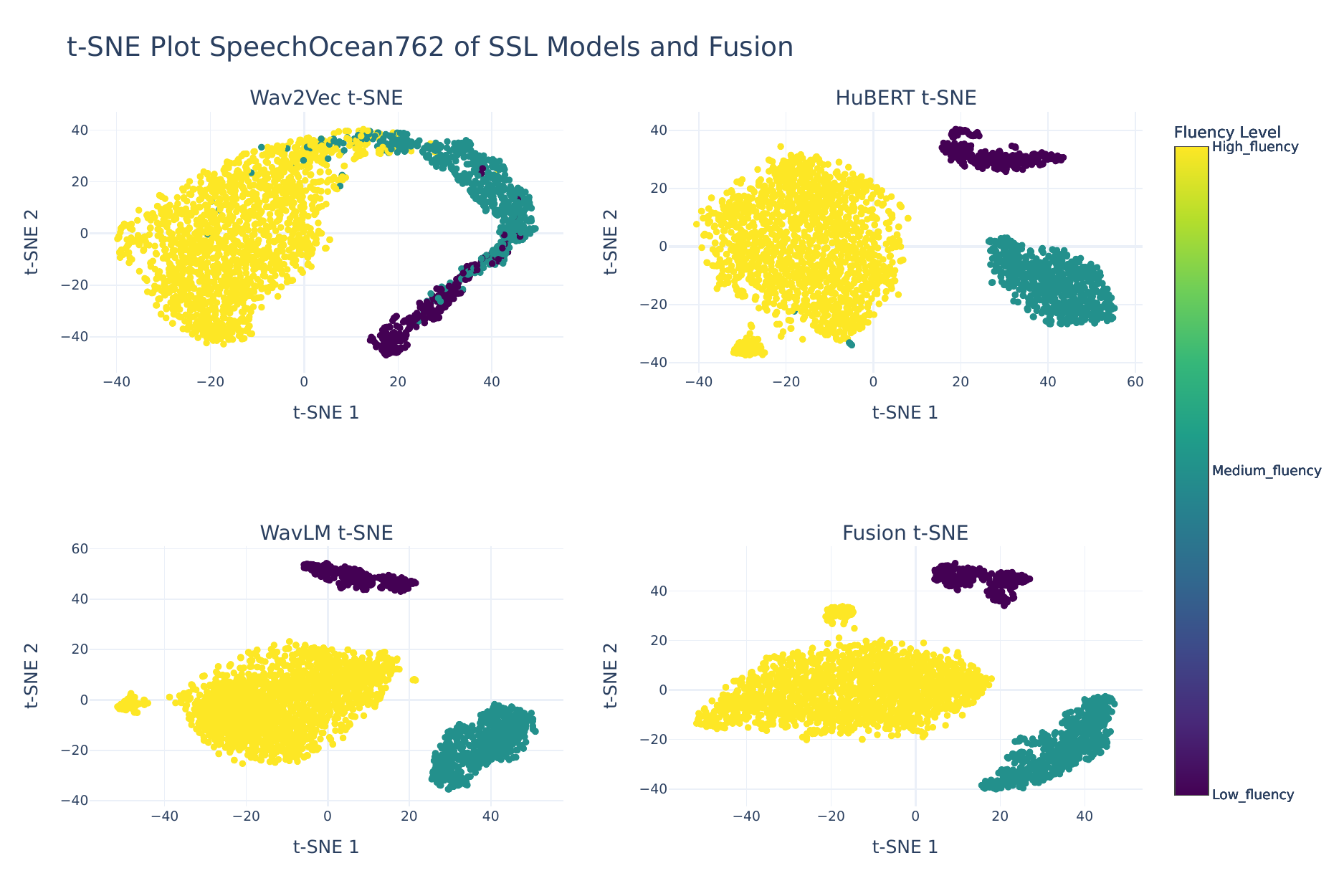}
   \caption{t-SNE Visualization of SSL embeddings (Speechocean762)}
    \label{fig:tsnet_spo762}
\end{figure}

To further analyze the effectiveness of SSL embeddings and fusion, we visualize their t-SNE projections for both datasets in Figures~\ref{fig:tsnet_avalinguo} and ~\ref{fig:tsnet_spo762}. These plots illustrate how well fluency levels are separated in the feature space. 

\paragraph*{Key Findings from t-SNE Analysis:}
\begin{enumerate}
    \item Fusion-based embeddings exhibit more coherent clustering compared to Wav2Vec2, and provide slight improvements in separability over HuBERT and WavLM.
    \item Fluency classes become more distinct with our approach, suggesting that chunk-based embeddings help disentangle fluency-related acoustic properties.
\end{enumerate}
These findings further support the use of fusion and chunking, as they lead to better feature space organization, which is crucial for fluency classification.

\subsubsection{Results and Discussion}
Our model exceeds baselines, with multi-SSL fusion enhancing phonetic and prosodic fluency representation. Silero-VAD-based chunking enhances segmentation, reducing noise while preserving speech rhythm.

Ablation results confirm that no single SSL model is sufficient; fusion consistently improves performance, with WavLM performing best among individual models. Analysis of fluency-related speech rate further supports the effectiveness of segmentation-aware models.

t-SNE visualizations reveal improved class separability with fusion, while voice quality analysis highlights instability patterns in non-fluent speech, suggesting that acoustic markers could enrich fluency assessment.

Our approach provides a scalable and adaptable fluency evaluation framework.

\section{Conclusion}
\label{sec:conclusion}
We introduced a chunk-based, multi-SSL fluency assessment framework, leveraging Wav2Vec2, HuBERT, and WavLM fusion with Silero-VAD segmentation. Our approach outperforms baselines and demonstrates that chunking enhances fluency modeling by preserving speech rhythm and disfluency markers.

Beyond SSL embeddings, our study suggests that voice quality metrics (F0, shimmer, HNR) can enrich fluency evaluation, bridging acoustic and linguistic fluency perspectives.

Future work could explore the adaptability of our approach to multilingual fluency assessment, particularly for low-resource languages such as Wolof, where fluency modeling remains underexplored. Other promising directions include integration with speech recognition models and attention-based fusion mechanisms for more dynamic weighting of SSL embeddings across linguistic contexts. Our approach assumes breath groups align with fluency, but this may not hold for speakers with irregular breathing patterns (e.g., due to anxiety). Future work could explore adaptive thresholding.


\printbibliography{}

\end{document}